%% file: eacl2024.tex
\pdfoutput=1

\documentclass[11pt]{article}

\usepackage{EACL2024}

\usepackage{times}
\usepackage{latexsym}
\usepackage{caption}
\usepackage{subcaption}
\usepackage{enumitem}
\usepackage[normalem]{ulem}
\usepackage{csquotes}

\usepackage[T1]{fontenc}

\usepackage[utf8]{inputenc}

\usepackage{microtype}

\usepackage{inconsolata}

\usepackage{graphicx}
\input{emnlp2023-latex/maths_command}
\usepackage{cleveref}
\usepackage{tikz}
\usepackage{tablefootnote}
 \usepackage{booktabs}

\makeatletter
\newcommand\footnoteref[1]{\protected@xdef\@thefnmark{\ref{#1}}\@footnotemark}
\makeatother

\usepackage{amsmath}

\title{LOCOST: State-Space Models for Long Document Abstractive Summarization}

\author{Florian Le Bronnec$^{*,1,2,3}$, Song Duong$^{*,1,6}$, Mathieu Ravaut$^{3,4}$, Alexandre Allauzen$^2$,\\ \textbf{Nancy F.\ Chen$^{3}$, Vincent Guigue$^{5}$, Alberto Lumbreras$^6$, Laure Soulier$^1$, Patrick Gallinari$^{1,6}$}\\ \vspace{-0.1cm}
\small$^1$Sorbonne Université, CNRS, ISIR, F-75005 Paris, France \\ \vspace{-0.10cm}
\small$^2$Miles Team, LAMSADE, Université Paris-Dauphine, Université PSL, CNRS, 75016 Paris, France \\ \vspace{-0.10cm}
\small$^3$Institute of Infocomm Research (I2R), A-STAR, Singapore \\ \vspace{-0.10cm}
\small$^4$Nanyang Technological University, Singapore  \\ \vspace{-0.10cm}
\small$^5$AgroParisTech, UMR MIA-PS, Palaiseau, France \\
\vspace{-0.10cm}
\small$^6$Criteo AI Lab, Paris, France \\ 
}
\begin{document}
\maketitle
\def\thefootnote{*}\footnotetext{Authors contributed equally to this work. Corresponding authors: \texttt{florian.le-bronnec@dauphine.psl.eu, s.duong@criteo.com}}
\begin{abstract}


State-space models are a low-complexity alternative to transformers for encoding long sequences and capturing long-term dependencies. We propose LOCOST: an encoder-decoder architecture based on state-space models for conditional text generation with long context inputs. With a computational complexity of $\mathcal{O}(L \log L)$, this architecture can handle significantly longer sequences than state-of-the-art models that are based on sparse attention patterns. We evaluate our model on a series of long document abstractive summarization tasks. The model reaches a performance level that is 93-96\% comparable to the top-performing sparse transformers of the same size while saving up to 50\% memory during training and up to 87\% during inference. Additionally, LOCOST effectively handles inputs exceeding \emph{600K} tokens at inference time, setting new state-of-the-art results on full-book summarization and opening new perspectives for long input processing.

\end{abstract}

\section{Introduction}

\input{emnlp2023-latex/0_introduction}

\section{Related Work}

\input{emnlp2023-latex/1_related_works}

\section{Background}\label{sec:background}

\input{emnlp2023-latex/2_background}

\section{Model}\label{sec:model}
\input{emnlp2023-latex/3_model}

\section{Experiments}\label{sec:experiments}

\input{emnlp2023-latex/4_experiments}

\section{Conclusion}\label{sec:conclusion}

Our paper explores a new encoder-decoder architecture dedicated to handle long input texts. By replacing the self-attention block by SSMs, we design a low complexity and lightweight model able to process long sequences up to 600K tokens at inference time on a single GPU. Our model achieves competitive results on summarization datasets. Moreover, surpassing the limits of existing sparse transformer alternatives, new state-of-the-art results are obtained on the BookSum-Book dataset. To the best of our knowledge, LOCOST is the first model able to process entire books without truncation, all in a single pass. These results offer exciting possibilities for abstractive text-processing tasks requiring extra-long sequences.



\section{Limitations}
\input{emnlp2023-latex/5_limitations}

\bibliography{LongSSM.bib}
\bibliographystyle{acl_natbib}

\appendix

\include{emnlp2023-latex/7_appendix}

\end{document}

%% file: emnlp2023-latex/maths_command.tex
\usepackage{amsmath,amsfonts,bm,mathtools}

\usepackage{tikz}
\usepackage{amssymb}
\usepackage{pifont}

\def\1{\bm{1}}

\def\vb{{\bm{b}}}
\def\vc{{\bm{c}}}
\def\vd{{\bm{d}}}

\def\vu{{\bm{u}}}

\def\vx{{\bm{x}}}
\def\vy{{\bm{y}}}

\def\mA{{\bm{A}}}

\def\mQ{{\bm{Q}}}

\def\mU{{\bm{U}}}
\def\mV{{\bm{V}}}

\newcommand{\bigO}{\mathcal{O}}

\newcommand{\R}{\mathbb{R}}
\newcommand{\C}{\mathbb{C}}

\DeclareMathOperator*{\ssm}{SSM}
\DeclareMathOperator*{\bissm}{BiSSM}

\newcommand*{\vkappa}{\boldsymbol{\kappa}}

\newcommand{\name}{LOCOST\,}

\newcommand{\vla}{\boldsymbol{\lambda}}
\DeclareMathOperator{\diag}{diag}
\DeclarePairedDelimiter\abs{\lvert}{\rvert}%
\DeclareMathOperator*{\argtopk}{arg\,top-K}

%% file: emnlp2023-latex/0_introduction.tex
Nowadays the design of efficient models for long texts remains an open challenge despite the recent progress achieved in natural language processing (NLP).
The introduction of transformer architectures \citep{attention_is_all_you_need} indeed came as a major bump in performance and scalability for text generation. However the quadratic complexity in the input length still  restricts the application of large pre-trained models to long texts. For instance, BERT \citep{bert} and BART \citep{bart} are limited to a context size of 512 and 1024 tokens respectively, which amounts to 2-3 paragraphs of standard text. 



\begin{figure}
    \centering
    \includegraphics[width=\linewidth]{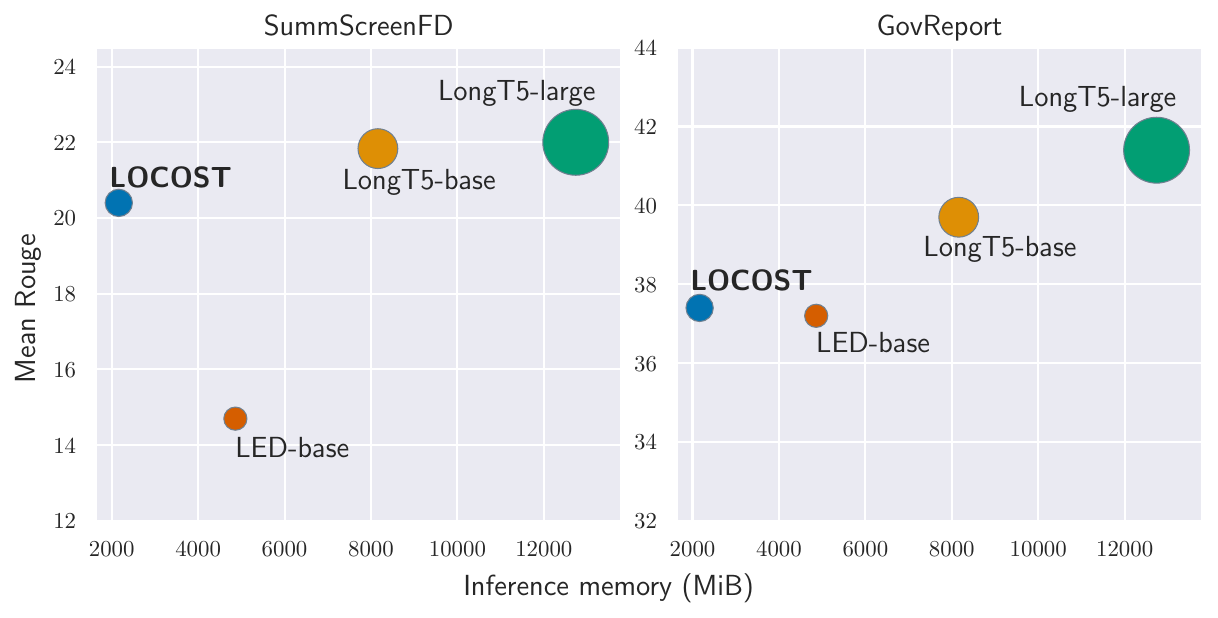}
    \caption{Mean ROUGE score with inference memory usage on long-document summarization with input length 16K (left: SummScreenFD dataset, right: GovReport dataset). The size of the circles represents the training memory usage. LOCOST demonstrates competitive performances compared to state-of-the-art sparse transformers of the same size, while being significantly more memory-efficient at both training and inference.
   }
   \vspace{-1.0em}
    \label{fig:scatter}
\end{figure}

To mitigate this issue, a straightforward approach is to leverage sparse-attention patterns \citep{sparse-transformer} to better cope with long texts. As key examples, \citet{longt5} and \citet{bigbird} extended the context capacity of  encoder-decoder models \citep{t5, pegasus} and showed drastic increases in the performance on long text summarization, motivating the quest to incorporate longer contexts.
However, in practice, even the best sparse-transformers need heavy computational resources to handle sequences of length larger than 8K tokens (see \Cref{fig:memory-exp}). 

Deep state-space models (SSMs) \citep{s4} have been proposed for sequence processing, with complexity $\bigO(L \log L)$, initially for computer vision and audio and more recently for text. Their recurrent architectures are designed for capturing long-range dependencies \citep{hippo}. 
Up to now, their applications have been restrained to either unconditional autoregressive generation, i.e., with a decoder-only \citep{h3,sashimi} ; or sequence classification, i.e., with an encoder-only \citep{s4,s4d,s4nd}. Tackling conditional text generation with SSMs as required e.g. for summarization remains yet unexplored.

In this paper, we propose LOCOST an encoder-decoder architecture to explore the performance of SSMs for conditional text generation tasks, through the lens of abstractive summarization. We demonstrate that SSMs can be competitive with transformer-based models while drastically reducing their memory requirements. We opt for a \emph{lightweight} architecture design, comparable to the average base transformers (roughly 250M parameters) in order to process extremely long sequences on standard compute resources. Our experimentations with extremely long sequences yield state-of-the-art results on the challenging BookSum-Book. With an increase of up to 2 points in average ROUGE score compared to sparse attention baselines, our model is able  to process entire books, without truncation, and on a single GPU. Our contributions are threefold:
\begin{itemize}[leftmargin=*]
    \item We propose a new encoder-decoder architecture based on state-space models. By bypassing the self-attention mechanism used in transformers, the model enjoys a complexity of $\bigO(L\log L$) instead of $\bigO(L^2)$ as in traditional transformers. 
    
    \item Compared with the best-performing sparse transformers of the same size, the model achieves 93-96\% of the best performance on various long document abstractive summarization while being up to 50\% more memory-efficient during training and up to 87\% at inference time,
    see \Cref{fig:scatter}. 
    \item The model is able to process entire input sequences of up to \emph{\textbf{600K tokens}}, a length far out of reach for sparse transformers. This allows the model to achieve a new state-of-the-art on a challenging full-book summarization task.

\end{itemize}
To the best of our knowledge, this is the first encoder-decoder that performs competitively with sparse transformers with no attention in the encoder. Furthermore, this work represents the first successful attempt at processing extremely long texts e.g. entire books without any truncation, all in a single pass. The proposed model opens new perspectives for addressing long texts with lesser resources.\footnote{Code and checkpoints available at \url{https://github.com/flbbb/locost-summarization}.} 


%% file: emnlp2023-latex/1_related_works.tex
In this section, we first review memory-efficient transformers and existing alternatives to the attention mechanism. Then, we discuss recent literature on state-space models.

\paragraph{Memory efficiency for transformers.} Reducing the memory consumption of transformers is an active research field. Optimization at the hardware level \citep{flash-attention} helped to improve the scaling of the attention computation on recent GPUs.
A line of work considers retrieving-augmented transformers, like \cite{retro,longmem}, that use additional modules to enhance the language modeling backbone. 
While crucial in developing memory-efficient architectures, we consider these last two topics as being orthogonal to our work that focuses on the models' architecture. Profuse literature focuses on tailoring the models' architecture for long inputs. Since the computational complexity of attention comes from the computation of the self-attention matrix, a straightforward way to reduce its cost is to approximate it using sparse-attention patterns. These patterns typically incorporate a combination of local attention and a set of carefully selected tokens. For instance, in addition to global tokens, BigBird \citep{bigbird} considers random tokens, while LSG \citep{lsg} considers sparse tokens through various strategy of sparsification. LongT5 \citep{longt5} chunks the sequence into blocks and averages their representations, which gives a number of global tokens equal to the number of blocks. An overview of the complexity of various sparse-transformers can be found in \Cref{tab:complexity}.

In contrast, we propose an alternative, computationally efficient architecture, without the need of costly self-attention blocks nor sparse-attention patterns.

\paragraph{Attention-free transformers.}
Some variants of transformers already avoid the standard attention mechanism. For example \citet{linear_transformer,transformer-quality} approximate the softmax similarity in the attention by a more efficient computation. 
More recently, mixing architectures were introduced in \citep{gmlp}. They are the main component of the FNet \citep{fnet} model, an encoder that replaces self-attention with a Discrete Fourier Transform (DFT). FNet has a complexity of $\bigO(L\log L)$ and is an encoder-only model, thus restricted to classification and regression tasks.

Our proposed model also bypasses attention in the encoder, reaching the same computational complexity as encoders such as FNet, while being a much more versatile model, specifically designed for conditional text generation.

\paragraph{State-space models (SSMs).} 

Deep learning implementations of SSMs consist of emerging architectures, first presented in \citep{hippo}. These architectures are particularly appealing for processing long sequences due to their reduced complexity compared to transformers, and their stronger theoretical guarantees compared to RNNs \citep{s4}, more details in \Cref{sec:background}. In practical applications, SSMs have found success in both classification and unconditional autoregressive generation for language modeling. \citet{s4} proposed a classification model that significantly improved the Long-Range Arena benchmark \citep{lra-benchmark}, which includes classification tasks involving images, synthetic sequences, and texts. Other studies have applied SSMs to video classification \cite{s4nd} and text classification \cite{pretraining-without-attention}. Regarding language modeling, many researchers have leveraged the natural causal formulation of SSMs, employing a decoder-only architecture for tasks like audio generation \citep{sashimi} and, more recently, autoregressive language modeling \citep{h3}.

In this work, we tackle the more challenging task of conditional text generation and study the performance of SSMs, used as an encoder-decoder architecture, on long document abstractive summarization. With our proposed architecture, we demonstrate the abilities of our model to process input sequences of up to 600K tokens, while being competitive to sparse-transformers on long document abstractive summarization.

%% file: emnlp2023-latex/2_background.tex
\begin{table}[t]
\centering
\input{emnlp2023-latex/tables/complexity}
\caption{Computational complexity per encoder layer as a function of the input length $L$, the local window size $w$ (typically set to 256 tokens), the number of global tokens $g$, random tokens $r$, sparse tokens $s$ and the chunk size $c$. LOCOST has a much lower complexity than other sparse-attention baselines.}
\vspace{-5pt}
\label{tab:complexity}
\end{table}

For contextualization, we leverage state-space models instead of self-attention. Throughout the paper, $L$ denotes the sequence length, $H$ the embedding dimension and $N$ the dimension of the state-space hidden state (to be introduced in \Cref{sec:background}). Before delving into our model in \Cref{sec:model}, we describe below  
the main components of the state-space architecture and elaborate on their potential for long sequence processing.


\paragraph{State-space models.} For unidimensional inputs $\vu = (u_0, ..., u_{L-1}) \in \R^L$, deep SSMs \citep{s4} are based on the recurrent equation:
\begin{equation}\label{eq:state-space}
\left\{
    \begin{aligned}
        \vx_{j+1} &= \mA \vx_j + \vb u_{j+1},\\
        y_{j+1} &= \vc^\top \vx_{j+1} + du_{j+1},
    \end{aligned}\right.
\end{equation}
where $\vx_j$ is the SSM hidden state and $y_j$ the output of the SSM. The state matrix $\mA \in \R^{N\times N}$  carries and transforms the hidden state through the iterations along with $\vb \in \R^N$, $\vc \in \R^N$, and $d \in \R$ which are learned parameters.


\paragraph{State-space convolution.} By unrolling the recurrence above, the output sequence $\vy \in \R^L$ can be expressed as: $y_j = \sum_{l=0}^j \vc^\top\mA^{j-l}\vb u_l + du_j$, $\forall l \in \{1,...,L\}$.
Let $*$ denote the causal convolution operator (details about this operator are in \Cref{app:convolution}). Then, we can define a convolution kernel $\vkappa \in \R^L$ that depends on $\mA, \vb, \vc$. A $\ssm$ layer is therefore parametrized by $\mA$, $\vb$, $\vc$, $d$ through $\vkappa$ and its output is defined by $\vy$ as in the following equation: 
\begin{equation}\label{eq:ssm-conv}
    \left\{
    \begin{aligned}
        \vy &= \vkappa * \vu + d\vu,\\
        \vkappa &= \left(\vc^\top \vb, \vc^\top\mA\vb, \dots, \vc^\top\mA^{L-1}\vb\right).
    \end{aligned}
    \right.
\end{equation}

For multidimensional $\vu \in \R^{L \times H}$, we simply compute $H$ convolutions with one kernel $\vkappa_h$ for each dimension.

\paragraph{SSMs efficiency.} Due to the linear time-dependency between hidden states, as shown in \Cref{eq:state-space}, we can compute the whole output $\vy$ directly as a convolution, without iteration over the time dimension, as opposed to RNNs.
A naive implementation of \eqref{eq:ssm-conv} would incur a quadratic complexity in the input length $L$, matching the complexity of transformers and thus be prohibitive for long sequences. However, thanks to the FFT, this computation can be performed in $\bigO(L \log L)$ (see \Cref{app:convolution} for more details).




%% file: emnlp2023-latex/tables/complexity.tex
\small
\begin{tabular}{lc}
\hline Encoder architecture & Complexity per layer \\
\hline Transformer (full) & $\bigO(L^2)$ \\
LED & $\bigO(Lw)$  \\
BigBird & $\bigO(Lw + L(g + r))$ \\
LSG & $\bigO(Lw + L(g + s))$ \\
LongT5 (TGlobal) & $\bigO(Lw + L\left\lfloor L /c\right\rfloor)$ \\
\hline
\name & $\bm{\bigO(L \log(L))}$ \\
\hline

\end{tabular}

%% file: emnlp2023-latex/3_model.tex
In this section, we present the \name model. 
We first introduce the bidirectional deep state-space model, then show how to use it to enable global contextualization of the tokens. Then, we present the architecture of the LOCOST layer with an efficient contextualization that can be used as a drop-in replacement for the self-attention mechanism in transformers. 

\begin{figure}[t!]
\centering
    \begin{subfigure}[t]{0.75\linewidth}
    \centering
    \includegraphics[width=\linewidth]{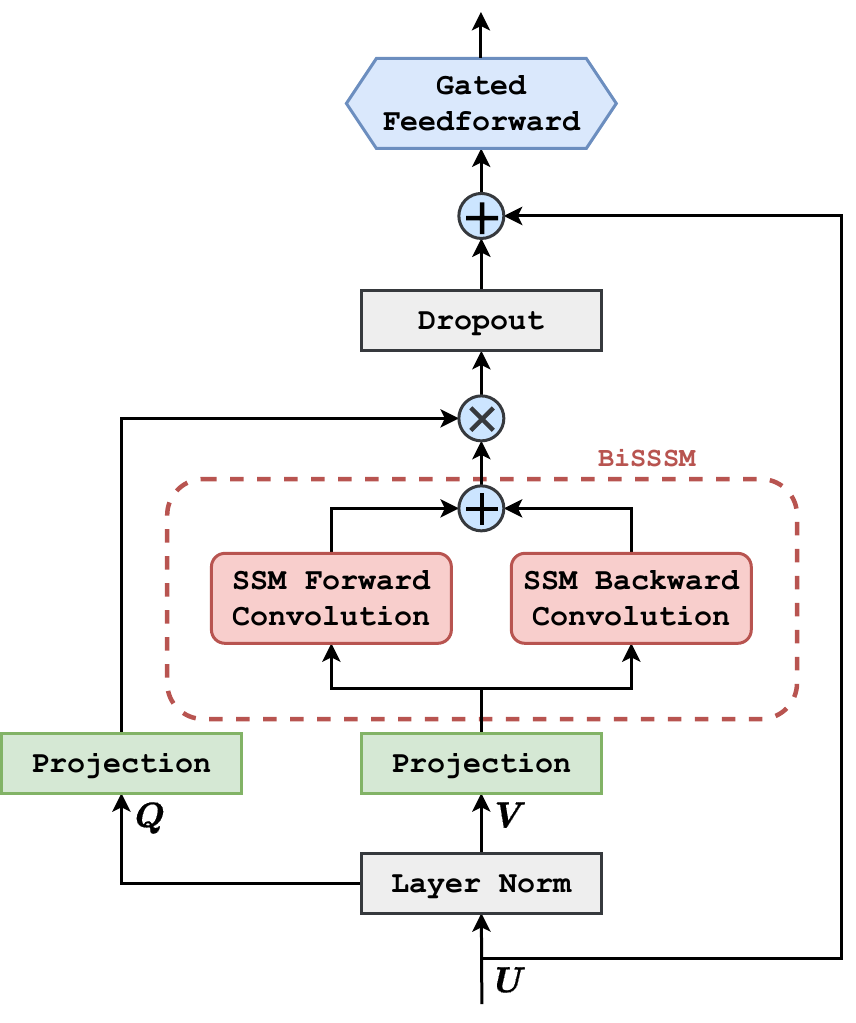}
    \caption{The \name layer.}
    \label{fig:archi-layer}
    \end{subfigure}
    \begin{subfigure}[t]{0.55\linewidth}
    \centering
   \includegraphics[width=\linewidth]{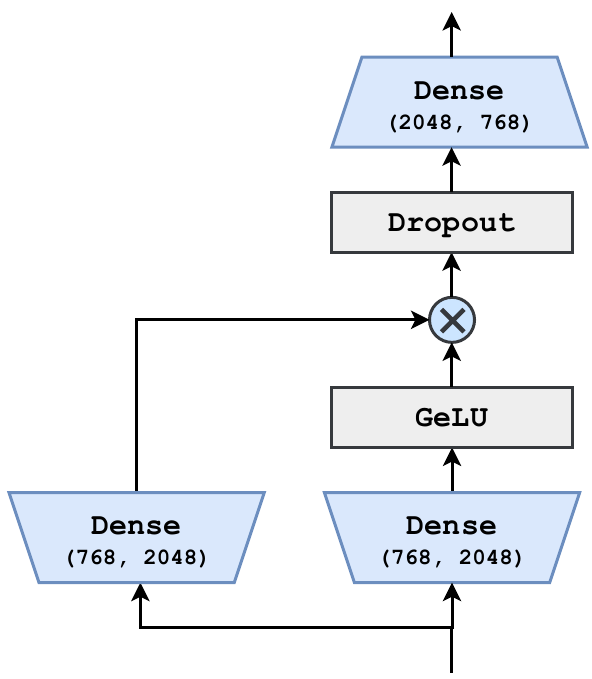}
   \caption{Gated feedforward net.}
   \label{fig:gated-ff}
    \end{subfigure}
    \caption{The embedded sequence is contextualized via a gated bidirectional SSM before passing through a gated feedforward net.}
    \label{fig:archi-model}
\end{figure}

\subsection{Capturing local and global contexts}

\paragraph{Intuition.} In deep SSMs, information from previous tokens flows up to the current token through the hidden states $\vx$. The convolution view provides another angle:  each output $\vy_j$ is a weighted sum of the previous tokens $\vu_0, \dots, \vu_j$, whose weights are given by $\vkappa$.

\paragraph{Bidirectional contextualization.} To aggregate information from both directions, we consider bidirectional convolutions. A first kernel,  $\overleftarrow{\vkappa}$ performs the regular causal convolution $\overleftarrow{\vkappa} * \vu$. A second kernel $\overrightarrow{\vkappa}$ is used to compute the cross-correlation with $\vu$.  The results of these two operations are summed out (similar to bi-recurrent encoder). The overall operation is described by the following equation:
\begin{align}\label{eq:bidirectional-conv}
    \vy_j &= \sum_{l \leq j} \overleftarrow{\vkappa}_{j-l} \odot \vu_l + \sum_{l \geq j} \overrightarrow{\vkappa}_{l-j}\odot \vu_l + \vd \odot \vu_j \nonumber \\
    &= \bissm(\mU)_j.
\end{align}
In this equation, $\mU \in \R^{L \times H}$ is the embedding matrix of the input text: $(\vu_0, \dots, \vu_{L-1})$. The kernels $\overrightarrow{\vkappa}, \overleftarrow{\vkappa}$ are computed as in \Cref{eq:ssm-conv}, with their respective parameters $(\overrightarrow{\mA}, \overrightarrow{\vc}, \overrightarrow{\vb})$ and $(\overleftarrow{\mA}, \overleftarrow{\vc}, \overleftarrow{\vb})$.
The element-wise product is denoted by $ \odot$ and we consider multidimensional inputs, with one kernel per dimension.

The output $\vy_j$ is now contextualized as a weighted sum of previous $\vu_{\leq j}$ and subsequent $\vu_{\geq j}$ inputs. For scalar inputs, more insights on how far in the future or in the past a scalar input $u_l$ contributes to the scalar output $y_j$ are given by the spectral radii
$\rho(\overrightarrow{\mA})$ and $\rho(\overleftarrow{\mA})$. Indeed the sensitivity of an output $y_j$ with respect to an input $u_l$ is bounded by the following quantity:
\begin{equation*}
\begin{split}
        \left|\dfrac{\partial y_j}{\partial u_l}\right| \leq
        \begin{cases}
            \rho(\overleftarrow{\mA})^{j-l} \abs{\overleftarrow{\vc}^\top\overleftarrow{\vb}} & \text{if} \quad l < j,\\
            \rho(\overrightarrow{\mA})^{l-j}\abs{\overrightarrow{\vc}^\top\overrightarrow{\vb}}  & \text{if} \quad l > j.
        \end{cases}
\end{split}
\end{equation*}


\begin{figure}
    \centering
    \includegraphics[width=.95\linewidth]{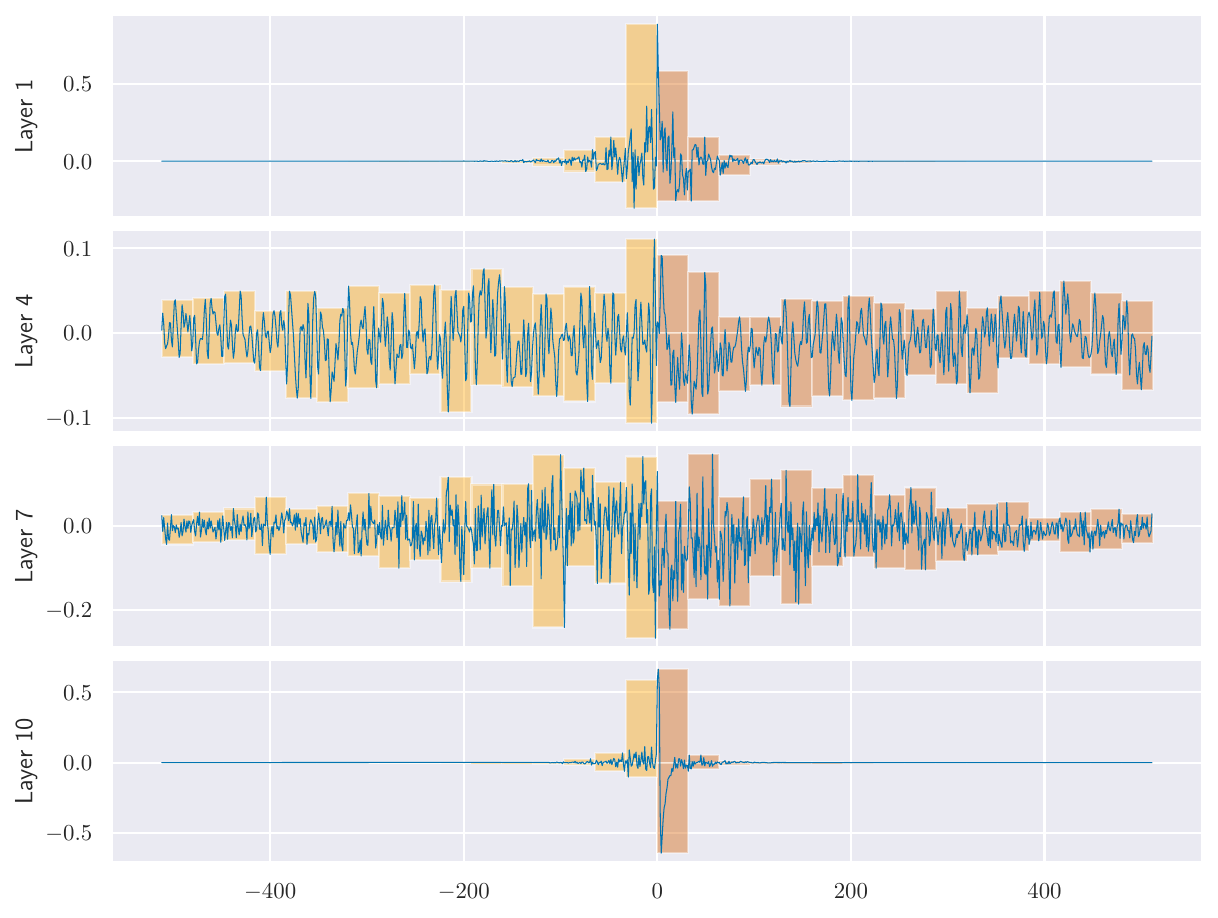}
    \caption{Visualization of the kernels corresponding to the first dimension for several layers of the pre-trained model. Bins show the average decay of the forward and backward kernels. This illustrates the different scales of each kernel. Layers 1 and 10 capture short and extra-short range contextualizations, while Layers 4 and 7 model extra-long and long contexts, respectively.
    }
    \label{fig:kernel_viz}
\end{figure}

For multidimensional inputs, using a state-space kernel for each dimension enables a fine-grained adjustment of the spectral radii independently for each of them. A small value corresponds to modeling local contexts, while a large value captures global ones. 

Some of the corresponding kernel weights of this convolution can be visualized on \Cref{fig:kernel_viz}. A more complete visualization can be found in \Cref{app:kernel_viz}.

\subsection{Architecture}
\paragraph{Encoder.}
Our encoder consists of a stack of \name layers, illustrated in \Cref{fig:archi-layer}. It is computed as follows:
\begin{itemize}[leftmargin=*]
    \item Embedding matrix $\mU \in \R^{L \times H}$ is first projected onto $\mQ, \mV \in \C^{L\times H}$.
    \item $\mV$ is contextualized through a $\bissm$.
    \item A pointwise multiplication $\mQ \odot \bissm(\mV)$ acts as a first gate before passing the output through a feedforward layer.
    \item This feedforward layer employs a second gating mechanism (see \Cref{fig:gated-ff}). For this component, we use gated GeLU that has shown to be efficient by \citet{glu}.
\end{itemize}
The architecture of the \name layer  (\Cref{fig:archi-layer}) resembles that of a transformer layer except that the self-attention mechanism is replaced by a gated bidirectional state-space model. We follow \citet{s4d} for the parametrization and initialization of the state-space models (more details in \Cref{app:ssm}).

\paragraph{Decoder.}
Since our focus is on long input summarization, the generation output length is very short compared to the input. For decoding, we follow the practice of other efficient architectures \citep{bigbird,longformer,longt5} and use a vanilla transformer decoder equipped with dense self- and cross-attention. A full description of hyperparameters of the model is provided in \Cref{app:hyperparameters}.

\paragraph{Complexity.} The \name layer takes $\bigO(H^2L + HNL + HL \log L)$ time and $\bigO(HNL)$ space to compute. We refer to \Cref{app:complexity} for more details.

%% file: emnlp2023-latex/4_experiments.tex
To validate our experiments, we focus on the long document abstractive summarization task as it represents a typical conditional generation problem with long input requirements.

\subsection{Experimental setup}
\paragraph{Approach.} We evaluate LOCOST following a classical pre-training then fine-tuning approach. For fine-tuning, we used the official train, validation and test splits of each dataset. We train all models until convergence and select the best model based on the validation Mean ROUGE (mean of ROUGE-1/2/LSum) for test evaluation.

\paragraph{Metrics.}
We evaluate LOCOST both with reference-based and reference-free metrics. For reference-based summarization evaluation, we use the traditional n-gram overlap summarization metrics \textbf{ROUGE-1/2/Lsum} \citep{rouge}. We average them into a single score to compare with other baselines. We also report \textbf{BERTScore} (BS) \citep{bertscore}, a model-based metric. For reference-free evaluation, we report the \textbf{BLANC} (BL) score \citep{blanc}, a metric that has been shown to correlate well with human evaluations. We also assess the throughput (samples per second) and the memory usage (MiB of GPU RAM) of LOCOST compared with other state-of-the-art sparse transformers.

\paragraph{Inference.}
In all of our experiments, we intentionally favored simplicity and opted for greedy decoding.

\subsection{Pre-training}

\paragraph{Pre-training objective.}
To pre-train the model, we leverage the gap-sentences generation (GSG) unsupervised pre-training objective, which was introduced by PEGASUS \citep{pegasus} and is well-suited for sequence-to-sequence generation. Unlike BART \citep{bart} or T5 \citep{t5} pre-training objectives, GSG endows the model with zero-shot summarization capabilities. GSG was successfully applied by subsequent generation models such as LongT5 \citep{longt5} and PEGASUS-X \citep{pegasus-x}. Namely, a document $D$ is split into its $M$ sentences: $D=\{s_1, \ldots, s_M\}$. Given a ratio $\alpha$, GSG then identifies $K=\lfloor \alpha M\rfloor$ sentences from $D$ that maximize the ROUGE-1 (noted $R\text{-}1$) with the rest of the document:
\begin{equation}
    U = \argtopk_j\: R\text{-}1\bigl(\bigcup_{i \neq j} \{s_i\}, s_j\bigr)
\end{equation}
The resulting subset $U \subseteq \{1, \ldots, M\}$ splits the document into a pseudo summary $\hat{Y}=\{s_i\}_{i \in U}$ and a pseudo-source $\hat{D}=\{s_i\}_{i \notin U}$, which are used for pre-training with the standard cross-entropy loss.

\paragraph{Pre-training data.}
We pre-train the model exclusively on the C4 dataset \citep{t5}, in BF16 for 1M steps, using an input sequence length of 4,096 and an output sequence length of 910.

\paragraph{Pre-training optimization.}
The learning rate scheduler we use is identical to T5, employing an inverse square root function, with the warm-up steps set to 10,000. We set the GSG-ratio $\alpha = 0.2$ and do not employ dropout during this phase. We follow closely the same pre-training as LongT5 \citep{longt5}.

\subsection{Fine-tuning}
\begin{table*}[h]
\centering
\resizebox{\textwidth}{!}{
\input{emnlp2023-latex/tables/summ_results}
}
\caption{Results on arXiv, PubMed and BookSum-Chapter with a input length of 4K, 4K and 8K tokens respectively. \% denotes the relative performance on the Mean ROUGE score w.r.t. LongT5, the best performing sparse-transformer at the given size, which is indicated as 100\%. BS stands for BERTScore and BL for BLANC.
}
\label{tab:summ_results}
\end{table*}

\begin{table*}[h]
\centering
\input{emnlp2023-latex/tables/SCROLLS_summ}
\caption{Results on the test set of SCROLLS for GovReport and SummScreenFD. L denotes the considered input length. \% denotes the relative performance on the Mean ROUGE score w.r.t. the reference LongT5. We reported baselines' results from the official SCROLLS test leaderboard. GovReport and SummScreen exhibit challenging long contexts sizes even for sparse transformers, as reported by the memory usage during training (MEM\textsubscript{train}) and inference (MEM\textsubscript{inf}) of the different architectures on 16K inputs. \mbox{\ding{55}} means out-of-memory.} 
\label{tab:scrolls_summ_results}
\end{table*}

\paragraph{Fine-tuning datasets.}
We evaluate LOCOST on a series of long-input abstractive summarization tasks. A table of statistics for all the datasets can be found in \Cref{app:dataset_stats}. 

\begin{itemize}[leftmargin=*,itemsep=3pt,parsep=3pt,topsep=3pt,partopsep=3pt]

\item \textbf{arXiv} \citep{arxiv-pubmed} Articles extracted from \emph{arXiv} using the core body document as the input sequence and the abstract as the target sequence.

\item \textbf{PubMed} \citep{arxiv-pubmed} Similar to arXiv, but articles come from \emph{PubMed}, a medical database.

\item \textbf{GovReport} \citep{gov-report} A long-document summarization dataset of US government reports with their executive summaries.

\item \textbf{SummScreenFD} \citep{summscreenfd} A long-document summarization dataset of TV series transcripts of entire episodes with human-written recaps of the episodes.

\item \textbf{BookSum (-Chapter \& -Book)} \citep{booksum} A collection of chapters from various books with a summary for each of them. We also consider the book-level version where the model has to summarize entire books. 

\end{itemize}

\paragraph{Fine-tuning optimization.}
We fine-tune in BF16 using a constant learning rate of $5\times 10^{-4}$ and a dropout rate of $0.1$ for all datasets. We experiment with lengths ranging from 4,096 to 32,768 for the input and 512 for the output, except for GovReport and BookSum-Book where we use 1024.

\paragraph{Baselines.} We consider both competitive sparse transformers, including LED \cite{longformer}, BigBird \cite{bigbird}, LongT5 \cite{longt5} and LSG \cite{lsg}, as well as dense encoder-decoders like BART \cite{bart}, T5 \cite{t5} and PEGASUS \cite{pegasus}.
For a fair comparison, \textbf{we only compare to sparse transformers architectures of equivalent size (roughly 250M parameters)}.

\subsection{Results}

\paragraph{Long-input summarization.}
\Cref{tab:summ_results} and \ref{tab:scrolls_summ_results} present our experimental results. Across all datasets, LOCOST reaches up to $96\%$ of state-of-the-art Mean ROUGE while being up to 3 times more memory-efficient than the best model LongT5 during both training and inference for 16K long inputs, e.g. on GovReport or SummScreenFD. The model is also twice as efficient as the local-attention transformer LED and up to 17 times more efficient than dense transformer BART at inference time. LOCOST significantly improves Mean ROUGE over LED and BigBird on all datasets while performing competitively with respect to LSG.
On all datasets, the results for LongT5 and LED have been obtained by fine-tuning from pre-trained checkpoints, following recommended configurations in \cite{longt5} and \cite{longformer} respectively. The results for BigBird has been reported from the original paper. LSG results are obtained from evaluating the publicly fine-tuned checkpoints on arXiv and PubMed and from our fine-tuning on BookSum-Chapter.
GovReport and SummScreenFD results are reported from the SCROLLS test leaderboard \citep{scrolls}.

\begin{figure*}[t!]
\centering
    \begin{subfigure}[t]{0.45\linewidth}
    \centering
    \includegraphics[width=0.9\linewidth]{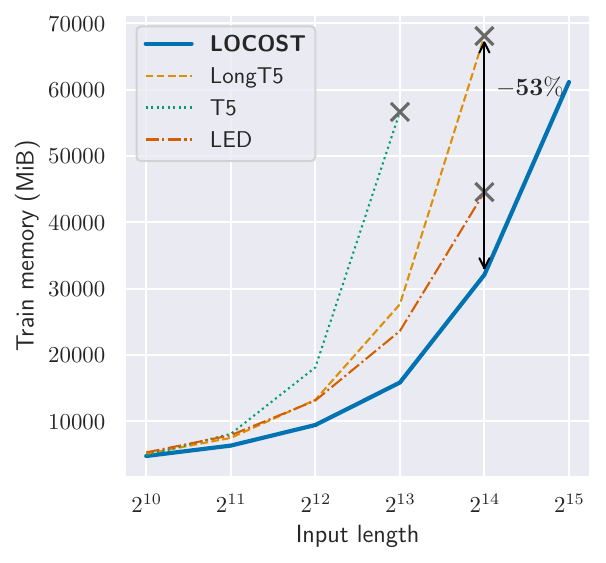}
    \end{subfigure}
    \begin{subfigure}[t]{0.45\linewidth}
    \centering
   \includegraphics[width=0.9\linewidth]{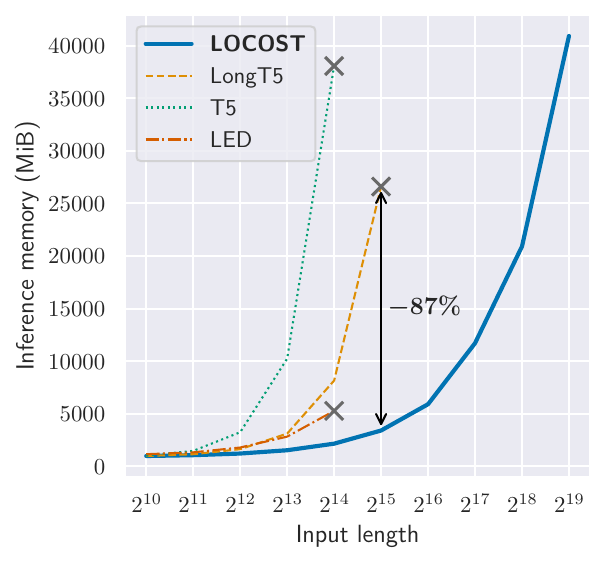}
    \end{subfigure}
    \caption{Memory consumption during a typical training (forward + backward) (\textbf{left}) and inference iteration (only forward) (\textbf{right}). Batch size = 1. Ending cross means out-of-memory or architectural limitations after this point.}
    \label{fig:memory-exp}
\end{figure*}

\paragraph{Throughput and Memory usage.} We measure the memory consumption of T5, LED, LongT5 and LOCOST on input lengths ranging from 1K to 500K tokens, at training and inference time. Results are presented on \Cref{fig:memory-exp}. Compared to LongT5, the best-performing baseline, LOCOST is able to process up to $2\times$ longer sequences during training and $16\times$ longer at inference time. This correlates also with a higher throughput during both training and inference, as shown in \Cref{tab:throughput_comparison}.



\begin{table*}
\centering
\input{emnlp2023-latex/tables/throughput}
\caption{Throughput comparison for different models at 4K and 16K input length.}
    \label{tab:throughput_comparison}
\end{table*}

\paragraph{Qualitative evaluation: GPT-3.5 preference.} Since our input texts are very long, performing a full human-based evaluation would be very costly and time consuming. Instead, we perform a mock human evaluation using GPT-3.5 \footnote{We use \textit{gpt-3.5-turbo-16k} model for evaluation.}. This practice has been used and has shown success in summary evaluation \citep{shen-llm-summarization,chatgpt-eval,chiang-lee-2023-large}. We ask the model to rate the generated summary on four dimensions: \textit{relevance}, \textit{consistency}, \textit{fluency}, and \textit{coherence}. More details are given in \Cref{app:gpt-eval}.

We perform evaluation on 500 samples randomly taken from PubMed. The results are shown in \Cref{tab:gpt_eval}. LOCOST produces summaries at a competitive level with respect to LongT5 (93-97\%).

\begin{table}[h]
\centering
\input{emnlp2023-latex/tables/gpt_eval}
\caption{GPT3.5 evaluation on PubMed with 4K input length using gpt-3.5-turbo-16k. \textbf{Rel} stands for \emph{relevance}, \textbf{Cons} for \emph{factual consistency}, \textbf{Flu} for \emph{fluency} and \textbf{Coh} for \emph{coherence}.}
\label{tab:gpt_eval}
\end{table}

\subsection{Extrapolating to longer sequences}

Because the lengths of the inputs considered during training are often limited due to complexity issues, a desirable property for a model would be to extrapolate at inference time to sequences much longer than the ones used during training. 

We train LOCOST on a maximum input length of 4,096 and evaluate it on the test set of arXiv with a maximum input length of 8,192 tokens. As shown in \Cref{tab:length_results}, this experiment confirms that LOCOST is indeed able to extrapolate to longer sequences than those employed in training. Note that LongT5 leverages relative positional encodings, enabling extrapolation capability. However, as previously mentioned, this comes at the expense of an increased complexity compared to LOCOST. 
In the next section, we push this idea further by considering extra-long sequences.



\begin{table}
\centering
\input{emnlp2023-latex/tables/length_gen_short}
\caption{Extrapolating to longer sequences experiments. L is the training sequence size. Gain represents the relative Mean ROUGE (Mean-R) improvement from evaluating on 4K to 8K maximum input length. The ROUGE increase asserts that both models are able to generalize to input lengths unseen during training.}
\label{tab:length_results}
\end{table}

\subsection{Extra-long sequences: towards full-book summarization}\label{sec:full-books-summ}
\begin{figure}
    \centering
    \includegraphics[width=0.85\linewidth]{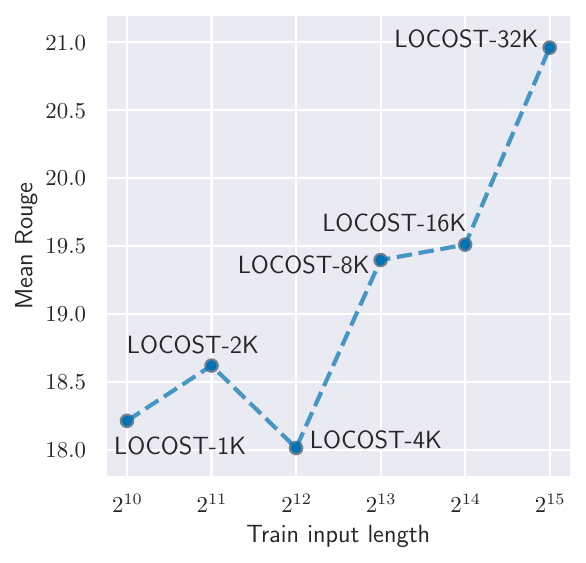}
    \caption{LOCOST trained on increasing sequence lengths evaluated on BookSum-Book dataset \emph{without truncation}, with texts reaching up to 600K tokens.}
    \label{fig:full_booksum_length}
\end{figure}

\paragraph{Effect of increasing contexts during training.} As shown previously, LOCOST exhibits a strong capability to generalize well on sequences longer than the ones seen during training. Due to the reduced memory usage at both train and inference time, we conduct in this section an analysis of its performances when facing extremely long texts e.g. \emph{summarizing entire books}. We consider the book-level setting of BookSum. We train multiple instances of LOCOST for 100 epochs on truncated books with a context length ranging from 1K to 32K and select the best model on Mean ROUGE on the validation set. We evaluate these models on the test set with untruncated books, and report the results in \Cref{fig:full_booksum_length}. We found that increasing the input length during training leads to an overall increase in the test Mean ROUGE scores as more contexts are being considered for optimization. Once more, this confirms the generalization capability of LOCOST on extra-long sequence lengths.

\begin{table}[t]
\centering\resizebox{\linewidth}{!}{
\input{emnlp2023-latex/tables/full_booksum}
}
\caption{Results on BookSum-Book. While being the smallest model, LOCOST achieves state-of-the-art on Mean ROUGE when summarizing entire books.}
\label{tab:full_booksum_results}
\vspace{-5pt}
\end{table}

\paragraph{Results on full-book summarization.} Based on the observations above, we put our best model LOCOST-32K to the test and compare it with LongT5 and current state-of-the-art models on BookSum-Book. For LongT5, we fine-tune the available checkpoint on the maximum possible input length during training (16K) and report its performance on the longest possible input length at inference time (32K). For the other models, the results come from the original papers, in which the models initially produce individual summaries for each paragraph of the book and then rank them according to the model's level of confidence. Results are shown in \Cref{tab:full_booksum_results}. Despite being the model with the least number of parameters, LOCOST achieves state-of-the-art Mean ROUGE compared to LongT5 and even large variants of BART, T5 and PEGASUS. LOCOST is also the only model capable of processing the full documents without truncation and handle sequence lengths of up to 600K tokens. This reveals that effectively processing full contexts without truncation can lead to strong performance enhancement.

%% file: emnlp2023-latex/tables/summ_results.tex
\small
\begin{tabular}{lccc|c|ccc|c|ccc|c}
\hline & \multicolumn{4}{c}{ \textbf{arXiv} } & \multicolumn{4}{c}{ \textbf{PubMed} } & \multicolumn{4}{c}{ \textbf{BookSum-Chapter} } \\
\textbf{Model} & R-1 / R-2 / R-L & BS & BL & \% &  R-1 / R-2 / R-L & BS & BL & \% & R-1 / R-2 / R-L & BS & BL & \%  \\
\hline PEGASUS\textsubscript{base} & 34.8 / 10.2 / 22.5 & -- & --  & 64.5 & 40.0 / 15.2 / 25.2 & -- & -- & 35.1 & --  & -- & -- & -- \\
\hline LED\textsubscript{base} & 41.3 / 16.4 / 37.7 & 60.1 & 10.6 &  91.3 & 42.9 / 19.6 / 39.7 & 63.1 & 15.9 & 89.2 & 17.3 / 2.6 / 15.7 & 41.7 & 4.4 & 46.3 \\
BigBird\textsubscript{base} & 41.2 / 16.4 / 37.0 & -- & -- & 90.4 & 43.7 / 19.3 / 39.9 & -- & -- & 89.8 & -- & -- & -- & -- \\
LSG\textsubscript{base} & 43.6 / 17.4 / 39.8 & 62.4 & 10.3  & 96.4 & 45.3 / 20.8 / 42.0 & 65.4 & 16.3 & 94.7  & 31.8 / 6.3 / 30.1 & 54.1 & 5.3 & 89.3 \\
LongT5\textsubscript{base} & 45.2 / 18.4 / 41.0 & 64.2 & 11.2  & 100 & 47.9 / 22.5 / 44.2 & 67.3 &  15.1 & 100  & 35.7 / 7.2 / 33.6 & 56.9 & 3.9 & 100 \\
\hline
LOCOST & 43.8 / 17.0 / 39.7 & 63.2 & 10.9 & 96.1 & 45.7 / 20.1 / 42.0 & 65.6 & 14.7 & 94.5 & 34.3 / 6.1 / 32.4 & 55.4 & 3.2 & 95.2 \\
\hline
\end{tabular}

%% file: emnlp2023-latex/tables/SCROLLS_summ.tex
\small
\begin{tabular}{lc|ccc|c|ccc|c|cc}
\hline & & \multicolumn{4}{c|}{ \textbf{GovReport} } & \multicolumn{4}{c|}{ \textbf{SummScreenFD} } & & \\
\textbf{Model} & L & R-1 & R-2 & R-L & \% & R-1 & R-2 & R-L & \% & MEM$_{\mathrm{train}}$ & MEM$_{\mathrm{inf}}$ \\
\hline BART\textsubscript{base} & 1K & 47.9 & 18.6 & 22.7 & 74.9 & 27.2 & 4.9 & 16.7 & 74.5 & \mbox{\ding{55}} & 17.6$\times$ \\
LED\textsubscript{base} & 16K & 56.2 & 26.6 & 28.8 & 93.7 & 24.2 & 4.5 & 15.4 & 67.3 & 1.0$\times$ & 2.3$\times$ \\
LongT5\textsubscript{base} & 16K & 57.7 & 30.0 & 31.4 & 100 & 34.8 & 9.6 & 21.1 & 100 & 2.9$\times$ & 3.8$\times$ \\
\hline
LOCOST & 16K & 56.5 & 26.8 & 28.9 & 94.2 & 33.4 & 8.1 & 19.7 & 93.5 & 1.4$\times$ & 1.0$\times$ \\
\hline
\end{tabular}

%% file: emnlp2023-latex/tables/throughput.tex
\small
\begin{tabular}{l|cc|cc}
    \hline & \multicolumn{2}{c}{ \textbf{Length 4K} } & \multicolumn{2}{c}{ \textbf{Length 16K} } \\
    \textbf{Model} & Inference (samples/s) & Training (samples/s) & Inference (samples/s) & Training (samples/s) \\
    \hline
    LED\textsubscript{base} & 3.57 & 1.69 & 1.67 & 0.45 \\
    T5\textsubscript{base} & 2.27 & 1.49 & 0.34 & \mbox{\ding{55}} \\
    LongT5\textsubscript{base} & 2.94 & 2.94 & 1.49 & 0.64 \\
        \hline
    LOCOST & 3.03 & 3.03 & 1.69 & 0.81 \\
        \hline
\end{tabular}

%% file: emnlp2023-latex/tables/gpt_eval.tex
\small
\begin{tabular}{lcccc}
\hline
\textbf{Model} & \textbf{Rel} & \textbf{Cons} & \textbf{Flu} & \textbf{Coh}  \\
\hline LongT5\textsubscript{base} & 4.6 & 4.7 & 3.7 & 3.7 \\
LOCOST & 4.3 & 4.4 & 3.6 & 3.5\\
\hline
\end{tabular}

%% file: emnlp2023-latex/tables/length_gen_short.tex
\small
\begin{tabular}{l|c|cc|c}
\hline & & \textbf{arXiv-4K} & \textbf{arXiv-8K}  \\
\textbf{Model}& L & Mean-R & Mean-R & Gain (\%)  \\
\hline LongT5\textsubscript{base}&4K & 34.8 & 35.5 & 2.0 \\
LOCOST & 4K & 33.5 & 34.3 &  2.4\\
\hline
\end{tabular}

%% file: emnlp2023-latex/tables/full_booksum.tex
\small
\begin{tabular}{lcccc|c}
\hline & & \multicolumn{4}{c}{ \textbf{BookSum-Book} } \\
\textbf{Model} & \#Params & R-1 & R-2 & R-L\tablefootnote{For a fair comparison with already existing results, we used ROUGE-L instead of ROUGE-Lsum on BookSum-Book.} & Mean-R \\

\hline 
BART\textsubscript{large} & 406M & 38.7 & 7.6 & 13.6 & 20.0 \\
T5\textsubscript{large} & 737M & \textbf{39.9} & 8.0 & 14.0 &  20.6 \\
PEGASUS\textsubscript{large} & 568M & 36.0 & 7.2 & 12.9 & 18.7 \\
LongT5\textsubscript{base} & 247M & 33.9 & 7.2 & 15.6 & 18.9 \\
\hline
LOCOST & 234M & 38.6 & \textbf{8.1} & \textbf{16.2} & \textbf{21.0}  \\
\hline
\end{tabular}

%% file: emnlp2023-latex/5_limitations.tex
Though we investigated lightweight models for computational reasons, scaling the architecture to a larger size could be studied. We focused on long document abstractive summarization, we leave for future work the study of SSMs on other long inputs abstractive tasks.
Although replacing self-attention with state-space encoders drastically reduces the computational complexity, the use of dense cross-attention in the decoder still limits the output sequence length in terms of computation during training.

\section{Ethics Statement}
We performed pre-training on a subset of the C4 dataset, which has been identified to include inappropriate content like hate speech and explicit material, as noted in the studies conducted by \citet{c4-undesirable} and also exhibits negative biases towards certain ethnicities \citep{documenting-c4}. It is important to investigate potential solutions for mitigating these problems through more meticulous preprocessing in order to prevent the emergence of such undesirable attributes in future research. Nevertheless, it is worth mentioning that despite these concerns, the C4 dataset serves as a benchmark within the community, and the reported results solely focus on the quality of the summaries, thereby avoiding any unethical implications.
In this paper, we consider a relatively small size for LOCOST. We believe our work could be reproducible with limited resources.
We tracked the GPU power consumption during pre-training. The average power usage was 190W per GPU. We trained for 140 hours on 16 GPUs. Given the local CO$_2$ intensity of 58 gCO$_2$/kWh \footnote{\url{https://www.eea.europa.eu/data-and-maps/daviz/co2-emission-intensity-13/}}, we can estimate that approximately 25kg of CO$_2$ have been emitted during the pre-training, to be compared with the average emissions of 4.6t of CO$_2$ par capita in 2019\footnote{\url{https://data.worldbank.org/indicator/EN.ATM.CO2E.PC}}.

\section{Acknowledgements}

This work has been partly funded through project ACDC ANR-21-CE23-0007 and ANR-23-PEIA-0008, PEPR IA, project "Principes théoriques et algorithmiques de l’apprentissage frugal (SHARP)".
This project was provided with computing AI and storage resources by GENCI at IDRIS thanks to the grants 20XX-AD011014060, 20XX-AD011014022 and 20XX-A0151014638 on the supercomputer Jean Zay's V100/A100 partition.
This work has also been partly funded through the Singapore International Pre-Graduate Award (SIPGA).

%% file: emnlp2023-latex/7_appendix.tex
\appendix
\newcommand{\vkappatilde}{\boldsymbol{\tilde{\kappa}}}
\newcommand{\vuhat}{\boldsymbol{\hat{u}}}
\newcommand{\vkappahat}{\boldsymbol{\hat{\kappa}}}
\section{Convolution}\label{app:convolution}
\subsection{Causal convolution}
In this section indices of sequence are represented by bracketed numbers.
The \emph{causal} convolution between sequences $\vu, \vkappa \in \R^{L}$ denoted as $*$ presented in \cref{sec:background} is defined as:
\begin{equation}\label{eq:causal-convolution}
    (\vkappa * \vu)[j] = \sum_{l=0}^{j} \kappa[j-l] u[l].
\end{equation}

\subsection{Convolution and DFT}
We are going to detail the link between convolution and the Discrete Fourier Transform. For that purpose, we need another tool, the \emph{circular convolution}.
\paragraph{Circular convolution.} Let's define $\vkappatilde$ the periodized version of $\vkappa$ as: $\forall j \in \mathbb{N},\ \tilde{\kappa}[j] = \kappa[j \bmod L]$.
For index $0 \leq j \leq L-1$, the discrete \emph{circular} convolution between $\vu$ and $\vkappa$ is defined as:
\begin{equation}\label{eq:circular-conv}
     (\vkappa \circledast \vu)[j] = \sum_{l=0}^{L-1} \tilde{\kappa}[j - l]u[l].
\end{equation}

\paragraph{Convolution theorem.} The convolution theorem states that (the derivation consists only in permuting the $\sum$ symbols):
\begin{equation}
    \vkappa \circledast \vu = \mathcal{F}^{-1} \left(\vkappahat \odot \vuhat\right),
\end{equation}

where $\boldsymbol{\hat{.}}$ designates the DFT of a sequence and $\odot$ designates the element-wise multiplication.

\paragraph{Causal convolution with DFT.} To compute $\vkappa * \vu$ with a DFT, a trick is to pad $\vkappa$ and $\vu$ with $L$ zeros \emph{before} taking their DFT. Indeed, if we replace $\vkappa$ and $\vu$ with their padded versions (hence vectors of $\R^{2L}$) in \cref{eq:circular-conv} we see immediately that it coincides with the \emph{causal} convolution \eqref{eq:causal-convolution}. This means that using the Fast Fourier Transform (FFT) algorithm, the causal convolution can be computed in $\bigO(L \log L)$.

\section{Hyperparameters}\label{app:hyperparameters}
The set of hyperparameters used are presented in \Cref{tab:hyperparameters}.
\begin{table}[h]
\centering
\input{emnlp2023-latex/tables/hyperparameters}
\caption{LOCOST hyperparameters.}
\label{tab:hyperparameters}
\end{table}

\section{Visualisation of learned kernels}
\begin{figure}[h]
    \centering
    \includegraphics[width=\linewidth]{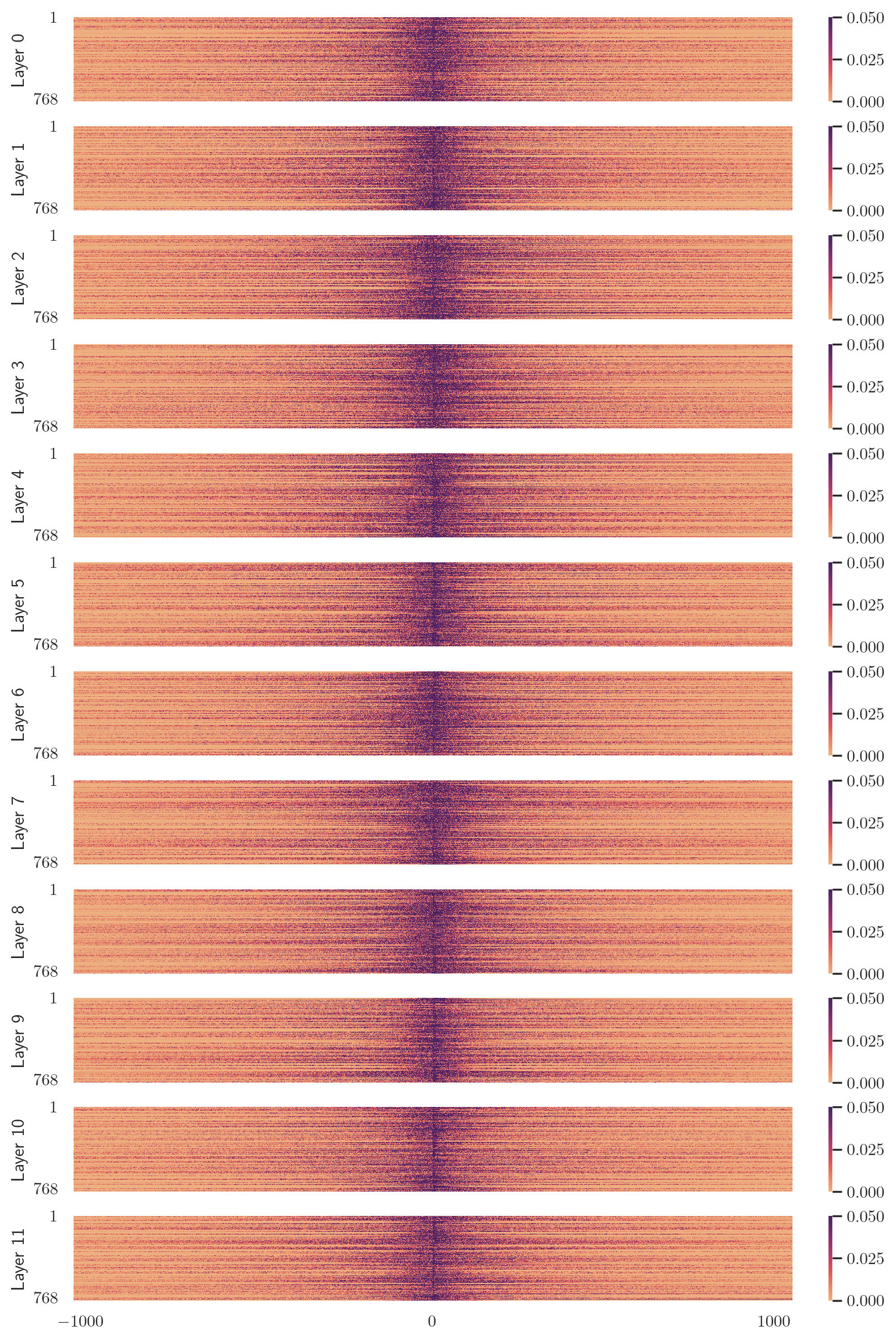}
    \caption{Visualization of the kernel (in absolute value) of size $768 \times 2048$ for each of the 12 layers. We clearly show that each layer has kernels of different scales that will model different context ranges.}
    \label{fig:imshow-kernels}
\end{figure}

A more complete visualization of the learned kernels can be found in \Cref{fig:kernel_viz} and \ref{fig:imshow-kernels}.
\label{app:kernel_viz}

\begin{figure}[h]
    \centering
    \includegraphics[width=\linewidth]{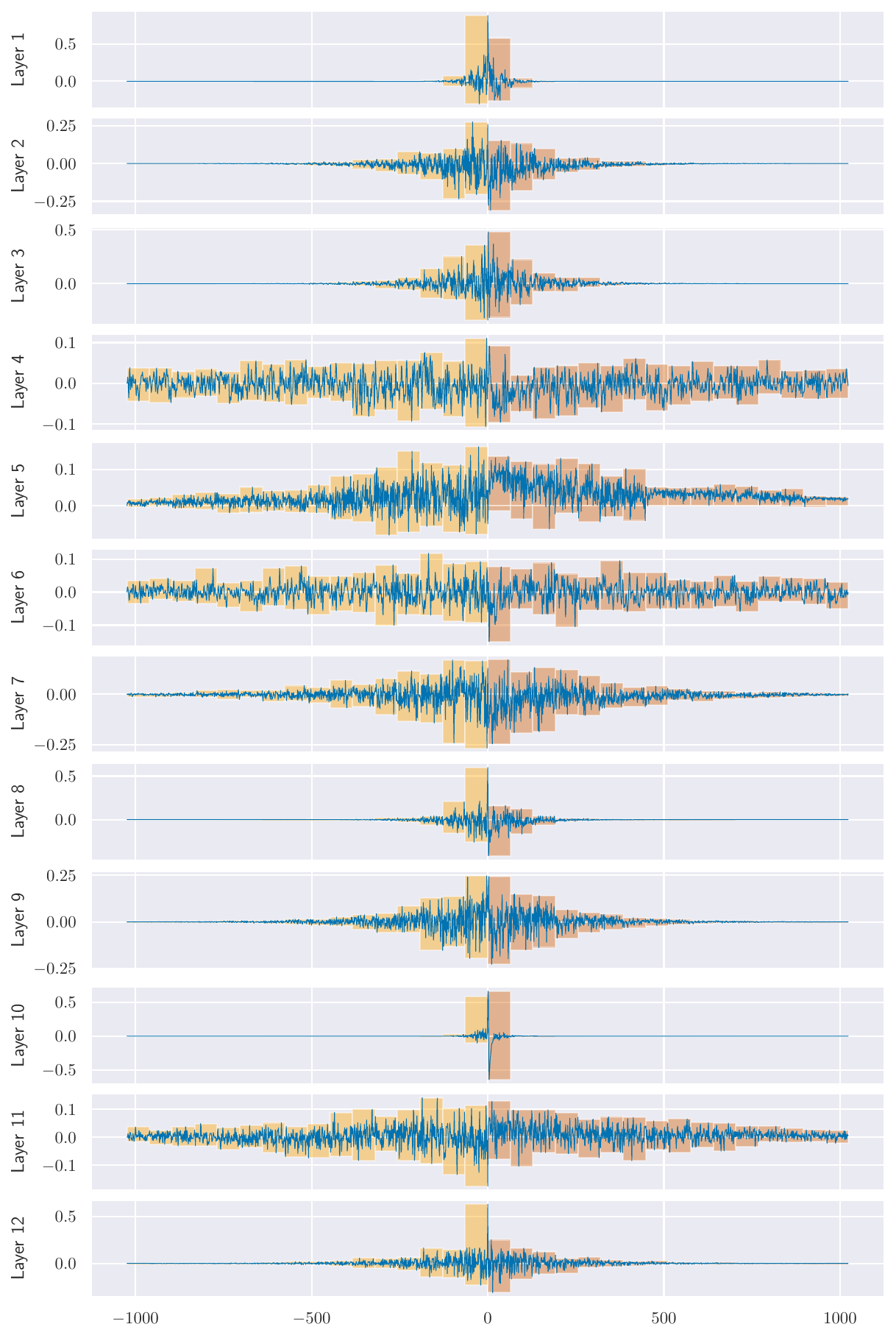}
    \caption{Complete visualization of the kernel of the first dimension of the model through all the 12 layers, includes visualization from \Cref{fig:kernel_viz}.}
    \label{fig:full_kernel_viz}
\end{figure}

\section{Computational complexity of a LOCOST layer}\label{app:complexity}
Projection onto $\mQ$ and $\mV$ takes $\bigO(LH^2)$ time and $\bigO(LH)$ space. Computing the SSM kernel $\vkappa = \left(\vc^\top \vb, \vc^\top\mA\vb, \dots, \vc^\top\mA^{L-1}\vb\right)$ takes $\bigO(LHN)$ time and space. Finally, calculating $H$ convolutions in parallel with DFT takes $\bigO(LH \log L)$ time.

\section{State-space models implementation details}
\label{app:ssm}
\paragraph{Parametrization.} We chose to follow the parametrization exposed in \citep{s4d}.
\begin{itemize}[leftmargin=*]
    \item The multi-dimensional state-tensor\footnote{\label{fn:complex}Using parameters in $\C$ gives better expressive power to the convolution, see \citet{s4d} for theoretical and empirical justifications.} $\mA \in \C^{H \times N\times N}$ is made of $H$ diagonal matrices $\mA_h = \diag(\vla_h) \in \C^{N \times N}$.
    \item For $0 \leq h \leq H$ and $0 \leq n \leq N$, $\vla \in \R^{H \times N}$ is $\lambda_{h, n} = \exp\left(\Delta_h\lambda_{h, n}^\mathrm{Re} + i\Delta_h\lambda_{h, n}^\mathrm{Im}\right)$.
    \item $\bm{\Delta} \in \R^H$ is a time-scaling parameter.
    \item We use $N = 256$. Most work chose either $N=64$ or $N=256$ \citep{s4d,h3}. Since increasing $N$ from $64$ to $256$ did only incur a negligible increase in memory consumption, we chose the latter, with the rationale that it should give more expressive power to $\vkappa$.
\end{itemize}

\paragraph{Initialization.} As reported in \citep{s4d} (see their Table~3), SSMs with special initialization are tailored for long inputs processing. This has been experimentally confirmed in \citep{spade}, where they use non-trainable state-space layers to provide long-range contextualization in addition to local attention.
\begin{itemize}[leftmargin=*]
    \item $\lambda_{h, n}^\mathrm{Re}$ is initialized to $-\dfrac{1}{2}$ and $\lambda_{h, n}^\mathrm{Im}$ to $\pi n$.
    \item $\Delta_h$ is initialized randomly following $\mathcal{U}([0, 1])$.
    \item $\vb, \vc \in \C^{N \times H}$ are initialized randomly following $\mathcal{N}(0, 1)$\footnoteref{fn:complex}. 
\end{itemize}

\section{Dataset details}\label{app:dataset_stats}
\paragraph{Statistics.} The statistics of the datasets can be found in \Cref{tab:dataset_stats}.
\begin{table*}[h]
\centering
\input{emnlp2023-latex/tables/datasets_statistics}
\caption{Statistics for the summarization datasets. Input length is computed using a SentencePiece tokenizer.}
\label{tab:dataset_stats}
\end{table*}

\paragraph{License.} C4: ODC-BY, arXiv/PubMed: unknown, BookSum: BSD-3-Clause, GovReport: unknown, SummScreenFD: unknown.

\paragraph{Usage.} All datasets were solely used for research purposes. Note that they are all in english and we refer to the original publications for more details.

\section{Implementation details}

\paragraph{Evaluation.} For ROUGE score computations, we used the implementation from \url{https://github.com/google-research/google-research/tree/master/rouge}, released under Apache 2.0 license. BERTScore was computed using the package \url{https://pypi.org/project/bert-score/} and is released under a MIT license. BLANC using \url{https://pypi.org/project/blanc/}, released under a MIT license.

\paragraph{Software.} Our code is based on Pytorch \citep{pytorch}, Huggingface \cite{huggingface-transformers} and H3 \cite{h3}. LongT5, LED models and weights are released under the Apache 2.0 license. The license for the LSG model and weights is unknown.

\section{Sample outputs}
Here is a sample summary (gold human abstract + generated summary) from PubMed.

\textbf{Ground Truth:} "Background : the purpose of this study was to perform the creating opportunities for parent empowerment ( cope ) program for iranian mothers and evaluate its effectiveness on stress, anxiety, and participation of mothers who have premature infants hospitalized in neonatal intensive care units ( nicus).materials and methods : a randomized clinical trial was conducted with 90 mothers of premature infants hospitalized in the educational neonatal nicus of state hospitals affiliated to isfahan university of medical sciences. For measuring the variables, state - trait anxiety inventory, the parental stressor scale : neonatal intensive care, and the index of parental participation / hospitalized infant were used. Intervention group received two phases of cope program. This program consisted of information and behavioral activities about the characteristics of premature infants. Sessions interval was from 2 to 4 days. Stress and anxiety were measured three times ( before each phase and 2 - 4 days after the second phase ). Mothers participation was evaluated 2 - 4 days after the second phase. The t - test, 2, mann - whitney u test, and repeated measurement test were used for data analysis. Results : mothers in the intervention group reported significantly less anxiety and less stress in the nicu after performing each phase of the cope program ( p 0.001 ), whereas at this time, the level of stress in the comparison group increased. Also, cope mothers participated in their infants' care rather than the mothers in the comparison group ( p 0.001). Conclusion : cope program was effective for iranian mothers. This study shows that irrespective of any culture, giving early and comprehensible educational - behavioral information may have positive effect on maternal psychological condition and maternal infant interaction."

\textbf{LOCOST:} "Background : the premature birth of an infant and the following neonatal intensive care cause psychological distress and can have a traumatizing effect on parents. The neonatal intensive care unit ( nicu ) environment has the potential to exacerbate stress for parents. The purpose of this study was to create opportunities for parent empowerment ( cope ) program in this survey.materials and methods : this is a quasi - experimental study. The study population consisted of 90 mothers with infants hospitalized in the nicu of isfahan, iran. The cope program included information about the behavior of infants, parenting role, and parenting stressor scales. Data were collected through demographic and infant variables questionnaire, state anxiety inventory, and parental stressor scale at the first contact point. Then, the cope program was given to the mothers. Data were analyzed by spss version 16 using descriptive statistics.results:the study resulted in the Absence of any significant difference in the state anxiety inventory and the parental stressor scale results. Moreover, the mean of maternal state anxiety inventory in the comparison group was significantly higher than the intervention group ( p 0.001 ). The parental stressor scale results showed significant difference between the two groups ( p 0.001 ). The state anxiety inventory showed significant difference between the two groups ( p 0.001 ). The parental stressor scale results showed significant difference between the two groups ( p 0.001). Conclusions : the cope program was feasible and the mother - infant positive interactions were high. Therefore, this study can be done by providing the parents with strategies that support them."

\section{GPT-3.5 evaluation}
\label{app:gpt-eval}
As a proxy for subjective assessment, we make use of \textit{gpt-3.5-turbo-16k} as a mock human evaluator. We ask the model to rate the generated summary on four dimensions: \textit{relevance}, \textit{consistency}, \textit{fluency}, and \textit{coherence} with the following prompt:
\begin{displayquote}
``Imagine you are a human annotator now. You will evaluate the quality of summaries written for an article. Please follow these steps: Carefully read the article, and be aware of the information it contains. Read the proposed summary. Rate the summary on four dimensions: relevance, consistency, fluency, and coherence. You should rate on a scale from 1 (worst) to 5 (best). Definitions are as follows:

Relevance: The rating measures how well the summary captures the key points of the article. Consider whether all and only the important aspects are contained in the summary.

Consistency: The rating measures whether the facts in the summary are consistent with the facts in the original article. Consider whether the summary does reproduce all facts accurately and does not make up untrue information.

Fluency: This rating measures the quality of individual sentences, whether they are well-written and grammatically correct. Consider the quality of individual sentences.

Coherence: The rating measures the quality of all sentences collectively, to fit together and sound natural. The article and the summary are given below:

Article: \textbf{\{insert article\}}

Summary: \textbf{\{insert summary\}}.

Rate the summary in the following format:

Relevance:

Consistency:

Fluency:

Coherence:''
\end{displayquote}

%% file: emnlp2023-latex/tables/hyperparameters.tex
\small
\begin{tabular}{lc}
\textbf{Parameter} & \textbf{Value} \\
\hline
Embedding dimensions $H$ & 768\\
Vocabulary size & 32100\\
Feedforward dimension & 2048\\
Activation function & GeLU\\
LayerNorm $\varepsilon$ & $1\times 10^{-6}$\\
State-space dimension $N$ & 256\\
Number of encoder layers & 12\\
Number of decoder layer & 12\\
Decoder attention heads & 12\\
\hline
AdamW $(\beta_1, \beta_2)$ & (0.9, 0.999)\\
AdamW weight decay & 0\\
Pre-training LR schedule & $\frac{2\times10^{4}}{\sqrt{\max(10^{4}, \, \text{current step)}}}$\\
Pre-training dropout & 0\\

Finetuning LR & $5\times 10^{-4}$\\
Finetuning LR schedule & constant\\
Finetuning dropout & 0.1
\end{tabular}

%% file: emnlp2023-latex/tables/datasets_statistics.tex
\small
\begin{tabular}{l|rrr|rrrr}
& \multicolumn{3}{c}{ \textbf{\#Examples per split} } & \multicolumn{4}{|c}{ \textbf{Input Length} } \\
\textbf{Dataset} & Train & Validation & Test & Average & Median & Max & 90$^{\mathrm{th}}$ \\
\hline arXiv & 203,037 & 6,436 & 6,440 & 10,720.18 & 8,519 & 378,825 & 20,170 \\
PubMed & 119,924 & 6,633 & 6,658 & 4,747.97 & 3,883 & 452,915 & 8,883 \\
GovReport & 17,457 & 972 & 973 & 10,576.06 & 8,840  & 240,734 & 18,834 \\
SummScreenFD & 3,673 & 338 & 337 & 9,589.36 & 9,044 & 26,447 & 15,171 \\
BookSum-Chapter & 9,600 & 1,484 & 1,431 & 5986.47 & 4311 & 204,567 & 11,804\\
BookSum-Book & 314 & 45 & 46 & 143,562.75 & 104,381 & 667,817 & 305,749 \\
\hline
\end{tabular}